\documentclass{article}

\usepackage{microtype}
\usepackage{graphicx}
\usepackage{subcaption}
\usepackage{booktabs} 
\usepackage[utf8]{inputenc}
\usepackage{tabularx}   
\usepackage[table]{xcolor}    
\usepackage{enumitem}  
\usepackage{xspace}
\usepackage{multirow}
\usepackage[colorinlistoftodos, prependcaption, textsize=tiny]{todonotes}
\usepackage{pgf} 
\usepackage{tikz} 
\usepackage[most]{tcolorbox}

\usepackage{makecell}
\usepackage[T1]{fontenc}
\usepackage{lmodern}

\newtcblisting{promptbox}[2][]{
  enhanced,
  breakable,
  colback=gray!3,
  colframe=black!35,
  boxrule=0.6pt,
  arc=2mm,
  left=6pt,right=6pt,top=6pt,bottom=6pt,
  listing only,
  title={#2},
  fonttitle=\bfseries,
  coltitle=black,
  attach title to upper,
  #1
}

\definecolor{slm_bg}{RGB}{245, 245, 245} 
\definecolor{llm_bg}{RGB}{235, 242, 255} 

\newcommand{\ours}{\textbf{DenseSteer}\xspace}

\newcommand{\rewrite}{\texttt{Dense-Rewriting}\xspace}
\newcommand{\reasoning}{\texttt{Dense Reasoning}\xspace}
\newcommand{\infam}{\textbf{InFamilySteer}\xspace}

\tcbset{
  aibox/.style={
    width=\linewidth,
    top=8pt,
    bottom=4pt,
    colback=blue!6!white,
    colframe=black,
    colbacktitle=black,
    enhanced,
    center,
    attach boxed title to top left={yshift=-0.1in,xshift=0.15in},
    boxed title style={boxrule=0pt,colframe=white,},
  }
}
\newtcolorbox{AIbox}[2][]{aibox,title=#2,#1}

\usepackage{hyperref}

\usepackage[accepted]{icml2026} 

\usepackage{amsmath}
\usepackage{amssymb}
\usepackage{mathtools}
\usepackage{amsthm}

\usepackage{fancyvrb}
\usepackage{fvextra}
\usepackage{spverbatim}
\usepackage{minted}

\usepackage[capitalize,noabbrev]{cleveref}

\theoremstyle{plain}

\theoremstyle{definition}

\theoremstyle{remark}

\usepackage[textsize=tiny]{todonotes}

\icmltitlerunning{DenseSteer: Steering Small Language Models towards Dense Math Reasoning}

\begin{document}

\twocolumn[
  \icmltitle{
DenseSteer: Steering Small Language Models towards Dense Math Reasoning 
}


  \icmlsetsymbol{equal}{*}

  \begin{icmlauthorlist}
    \icmlauthor{Yang Ouyang}{ncsu}
    \icmlauthor{Shuhang Lin}{rutgers}
    \icmlauthor{Jung-Eun Kim}{ncsu}
  \end{icmlauthorlist}

  \icmlaffiliation{ncsu}{North Carolina State University}
  
  \icmlaffiliation{rutgers}{Rutgers University}

  \icmlcorrespondingauthor{Jung-Eun Kim}{jung-eun.kim@ncsu.edu}

  \icmlkeywords{Math Reasoning, Large Language Models, Machine Learning, ICML}

  \vskip 0.3in
]



\printAffiliationsAndNotice{}  

\begin{abstract}
  Large language models (LLMs) demonstrate strong chain-of-thought (CoT) reasoning abilities, while smaller models ($\leq$ 3B parameters) significantly underperform on multi-step reasoning tasks.
  Based on empirical analyses of the Qwen-2.5 model family on math reasoning benchmarks, we find that more proficient reasoning is associated with fewer reasoning steps but higher information density per step, a property we term \textit{Dense Reasoning}. 
  Motivated by this observation, we propose \ours, a training-free inference-time steering framework that enhances small-model reasoning by modulating internal representations toward dense reasoning patterns. 
  Experiments show that our method yields consistent accuracy improvements without increasing token-level Negative Log-Likelihood, highlighting dense reasoning as an effective structural approach to mathematical problem solving. 
\end{abstract}

\begin{figure}[t]
    \centering
    \includegraphics[width=0.65\linewidth]{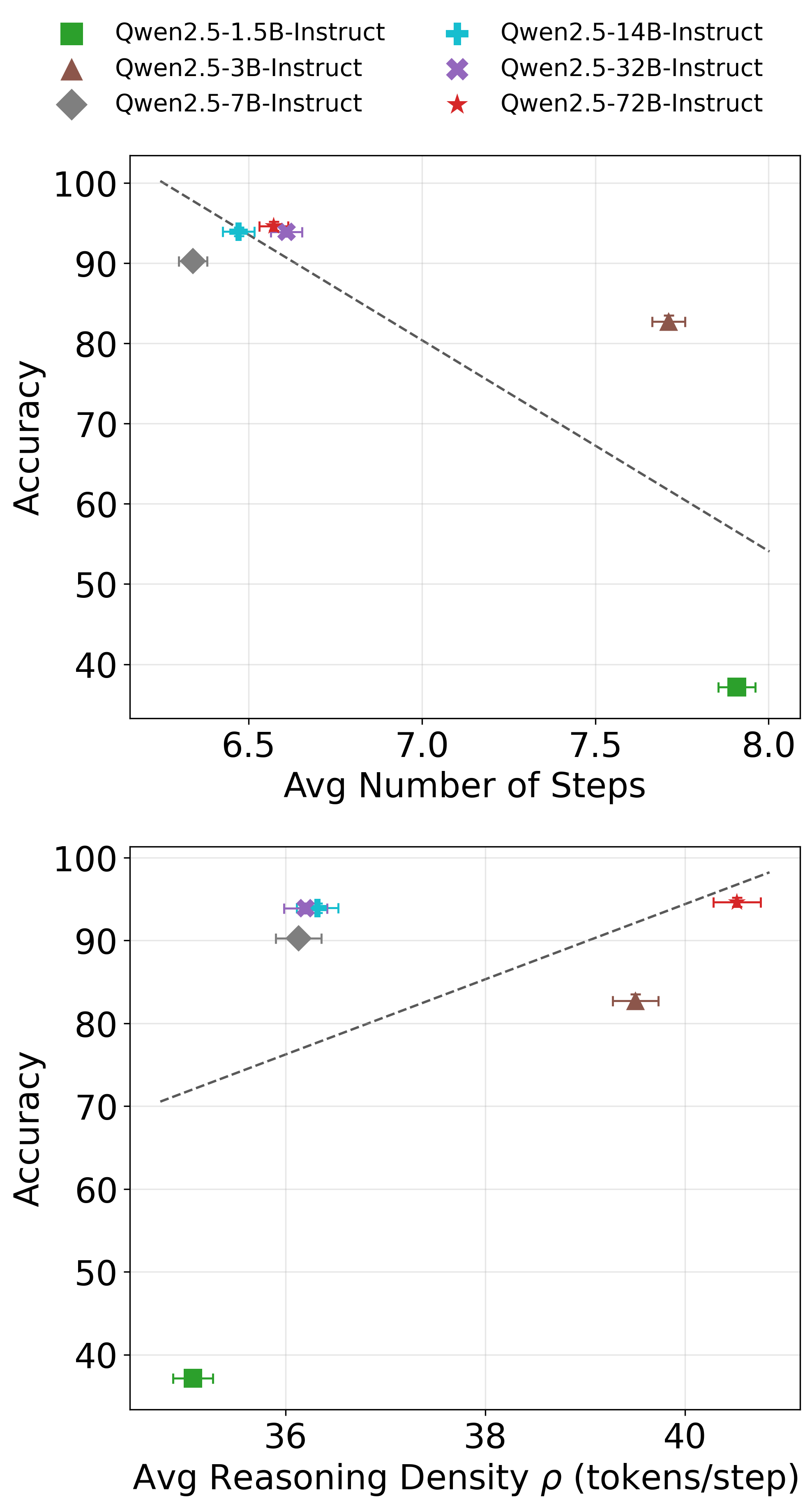}

\caption{\textbf{Analysis of Reasoning Patterns on GSM8K.}
For each model and each GSM8K test problem, we sample 8 responses.
Error bars denote $\pm 1$ standard error of the mean (SEM), computed over question-level averages.
Across Qwen2.5-Instruct models, stronger models often use fewer reasoning steps and higher per-step density than smaller, lower-accuracy models.
This coarse trend motivates our focus on dense, model-aligned reasoning.}
    \label{fig:stats_comparison}
\end{figure}

\begin{figure*}[t]
\vspace{-0.1in}
    \centering
    \includegraphics[width=0.95\textwidth]{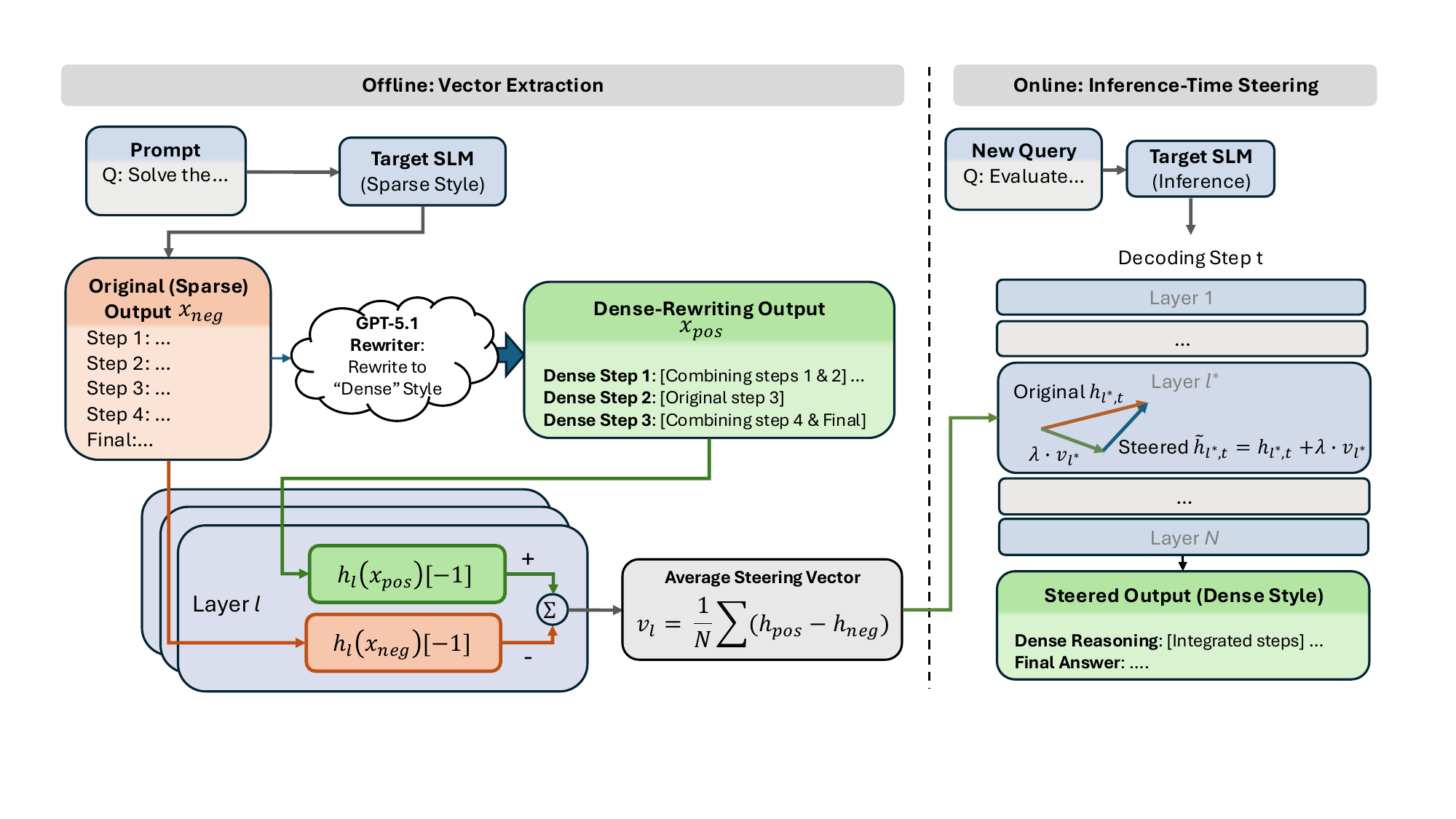}
    \vspace{-0.3in}
    \caption{ Overview of the proposed \ours framework. Given a query, the target model first produces a baseline reasoning trace. This trace is minimally rewritten into a dense variant while preserving semantic content, forming a contrastive pair.
Hidden-state differences between dense and sparse variants are aggregated to extract a steering vector at a selected layer.
At inference time, this steering vector is injected into the residual stream to guide the target model toward denser reasoning without modifying model parameters.
}
    \label{fig:03_method_overview}
\end{figure*}

\section{Introduction}

Large Language Models (LLMs) \citep{brown2020languagemodelsfewshotlearners} demonstrated remarkable problem-solving abilities, often by producing step-by-step reasoning before giving the final answer, which is known as Chain-of-Thought (CoT) \citep{ cobbe2021trainingverifierssolvemath, wei2022cot, wei2023chainofthoughtpromptingelicitsreasoning, cobbe2021trainingverifierssolvemath, shao2024deepseekmathpushinglimitsmathematical, yang2024qwen25mathtechnicalreportmathematical}. By structuring complex problem-solving into a coherent chain of rationales, this technique enhances performance and interpretability.

However, the practical utility of LLMs is hindered by their prohibitive deployment costs, a burden compounded by the additional token overhead inherent in CoT reasoning. To mitigate these limitations, recent research has pivoted toward Small Language Models (SLMs) that retaining a subset of LLM capabilities at reduced footprints. 
Despite their efficiency advantages, compact architectures (typically $\leq$ 3B parameters) exhibit a significant performance gap compared to frontier models \citep{anthropic2023claude, openai2023gpt4, touvron2023llamaopenefficientfoundation}. This gap persists even within the same model family (e.g., Qwen2.5-3B-Instruct vs. Qwen2.5-72B-Instruct).

One widely adopted approach to bridge this gap is Knowledge Distillation (KD) \citep{hinton2015distillingknowledgeneuralnetwork, gu2024minillmknowledgedistillationlarge, deepseekai2024deepseekv3technicalreport, agarwal2024onpolicydistillationlanguagemodels}, where the rationales generated by a larger ``teacher'' model are used to fine-tune a smaller ``student'' model. While effective, KD remains computationally expensive, requiring extensive training time, significant GPU resources, and large-scale teacher-generated datasets. More critically, small models often struggle to absorb strong teacher traces, leading to a learnability gap and severe distribution mismatch \citep{li2025small}.

In contrast, we explore a lightweight \textit{inference-time} alternative based on Steering Vectors~\citep{panickssery2024steeringllama2contrastive}, which modulate model behavior by shifting activation states requiring negligible computational overhead. Steering vectors are extremely data-efficient and can be constructed from as few as 50 contrastive samples. However, naively steering an SLM toward the raw hidden states of a much larger model results in severe distribution misalignment, manifesting as high Negative Log-Likelihood (NLL) as we empirically demonstrate in Section~\ref{subsec:nll_alignment}.

Our key motivational observation is a coarse-grained reasoning pattern discrepancy: higher-performing models often produce less fragmented traces, using fewer intermediate steps while maintaining higher information density per step, as shown in Figure~\ref{fig:stats_comparison}. We term this phenomenon \textbf{Dense Reasoning}, quantified by our proposed Reasoning Density metric introduced in Section~\ref{subsec:self_dense_guidance}. This asymmetric pattern motivates our hypothesis that dense reasoning is a useful structural target. We support this claim through interventions demonstrating that steering models toward denser reasoning improves both reasoning structure and task accuracy.

To transfer this structure without inducing distribution shift, we propose \ours \footnote{Our code is available at: \url{https://github.com/oyy2000/DenseSteer}}. Instead of using the large model's traces directly as positive samples, we utilize commercial model GPT-5.1 to rewrite the \textit{SLM's own outputs} to mimic the larger counterparts' ``Dense Reasoning'' style - we term this approach as \textbf{Dense-Rewriting}. This creates an in-domain positive anchor that aligns with the student’s distribution while adopting the superior reasoning structure. By constructing our steering vectors from these rewritten samples, \ours achieves a dual gain: it significantly boosts accuracy on math reasoning tasks by enforcing dense reasoning, while maintaining low NLL.

In summary, our contributions are:
\begin{itemize} [leftmargin=*, topsep=2pt, itemsep=2pt, parsep=0pt]
    \item We identify \textbf{Dense Reasoning} as a structural characteristic of proficient mathematical reasoning in large language models, where solutions are expressed with fewer intermediate steps but higher information \emph{density} per step.
    \item We propose \ours, a lightweight, training-free, inference-time paradigm that effectively enhances SLM performance with extremely few samples.
    \item We enable it through \textbf{Dense-Rewriting}, a rewrite-based contrastive pair construction method that preserves in-distribution generation while mitigating distribution shift, enabling stable accuracy improvements without increasing NLL.
\end{itemize}
\newpage
\section{Related Work}
\subsection{Knowledge Distillation and Training-Based Reasoning Enhancement}

Knowledge distillation has been widely adopted to transfer reasoning ability from large language models to smaller students using teacher-generated rationales or trajectories \citep{hinton2015distillingknowledgeneuralnetwork, gu2024minillmknowledgedistillationlarge, deepseekai2024deepseekv3technicalreport, agarwal2024onpolicydistillationlanguagemodels}.
Recent works demonstrate that structured supervision on reasoning traces can substantially improve mathematical reasoning in compact models.

Several studies further explore specialized distillation or training strategies for efficient reasoning.
Skip-Thinking \citep{chen2025skipthinking} introduces chunk-wise chain-of-thought distillation to enable faster and more accurate reasoning, while Phi-4-Mini-Reasoning \citep{xu2025phi4mini} investigates tailored training recipes for small reasoning models using distilled CoT data.
While effective, these approaches rely on training-intensive pipelines and large-scale teacher supervision.
Moreover, strong teacher traces may induce severe distribution mismatch for compact architectures, leading to the learnability gap \citep{li2025small}.
In contrast, our work focuses on inference-time transfer of reasoning structure without parameter updates.

\subsection{Inference-Time Steering and Representation-Level Control}

Inference-time intervention methods aim to modulate model behavior without retraining by manipulating hidden representations or decoding dynamics \citep{panickssery2024steeringllama2contrastive, turner2024activationaddition, zou2023representationengineering, hojerImprovingReasoningPerformanceLargeLanguage2025}.
These techniques have been applied to style control, safety alignment, and bias mitigation.

Recent studies suggest that representation-level interventions can also influence higher-level reasoning behaviors.
For example, SEAL calibrates explicit long-CoT reasoning traces by steering away from redundant reflections and transition patterns. \citep{chen2025seal}.

Other works analyze the internal structure of multi-step reasoning, highlighting the importance of stable intermediate representations and latent reasoning circuits \citep{wang2025understandinglanguagemodelsolve, yang2025EMNLP}.
Our work builds on this line of research by demonstrating that reasoning structure--specifically, reasoning density--can be effectively transferred via lightweight, inference-time steering, without auxiliary training or distribution shift.

Our work builds on this line of research, but differs in the construction and selection of the steering signal.
Most prior steering setups contrast behaviorally or semantically opposite examples, such as positive versus negative demonstrations, target versus opposite preferences, or safe versus unsafe prompts.
\ours instead rewrites the same solution into a denser version while preserving its reasoning semantics, and uses target-model NLL as an explicit compatibility criterion.
Thus, our contribution is an in-distribution contrastive signal for steering reasoning density without auxiliary training or large distribution shift.

\section{Preliminaries and Observations}
\label{sec:preliminary}
Inspired by prior findings in knowledge distillation \citep{son2021denselyguidedknowledgedistillation} and recent analyses of the \emph{learnability gap} in small language models \citep{li2025small}, we recognize that the effectiveness of knowledge transfer is fundamentally constrained by the distributional discrepancy between the target (student) model and the reference (teacher) model.
In particular, even when reference responses are correct or high-quality, they may be difficult for a small model to absorb if they lie far outside the student’s intrinsic generation distribution. 

Since our approach counts on \emph{positive samples} which in our setting could be responses from a reference model, we require a criterion to select positive samples that are distributionally compatible with the target model. We therefore introduce a simple yet effective measure to identify positive samples that minimize the distributional difference with respect to the target model, thereby improving the stability and effectiveness of inference-time steering.

\subsection{Quantifying Distribution Difference}
\label{subsec:nll_alignment}

To quantify how compatible the reference traces are with the target model, we evaluate how natural these traces appear under the student model by computing the token-level negative log-likelihood (NLL) on a shared subset of $N=200$ GSM8K samples.
Formally, for a reference rationale token sequence, $x_{1:T}$, we compute:
\begin{equation}
\mathrm{NLL}(x; P_{\theta})
= -\frac{1}{T} \sum_{t=1}^{T} \log P_{\theta}(x_t \mid x_{<t}),
\end{equation}
which measures the average token-level likelihood assigned by the target model to the reference tokens. Intuitively, a lower NLL indicates a smaller distribution difference and better alignment with the student model's intrinsic generation prior.

Using this metric, we compare the distribution difference between the student model $P_{\theta}$ and two candidate sources of reference samples:
\begin{enumerate}
    \item \textbf{Same-family reference samples:} reasoning traces generated by a larger model from the same family (e.g., Qwen2.5-7B-Instruct);
    \item \textbf{Cross-family reference samples:} reasoning traces generated by a strong but architecturally different model (e.g., Llama-3.2-8B-Instruct).
\end{enumerate}

\begin{figure}[t]
    \centering

    \begin{subfigure}[t]{0.9\linewidth}
        \centering
        \includegraphics[width=\linewidth]{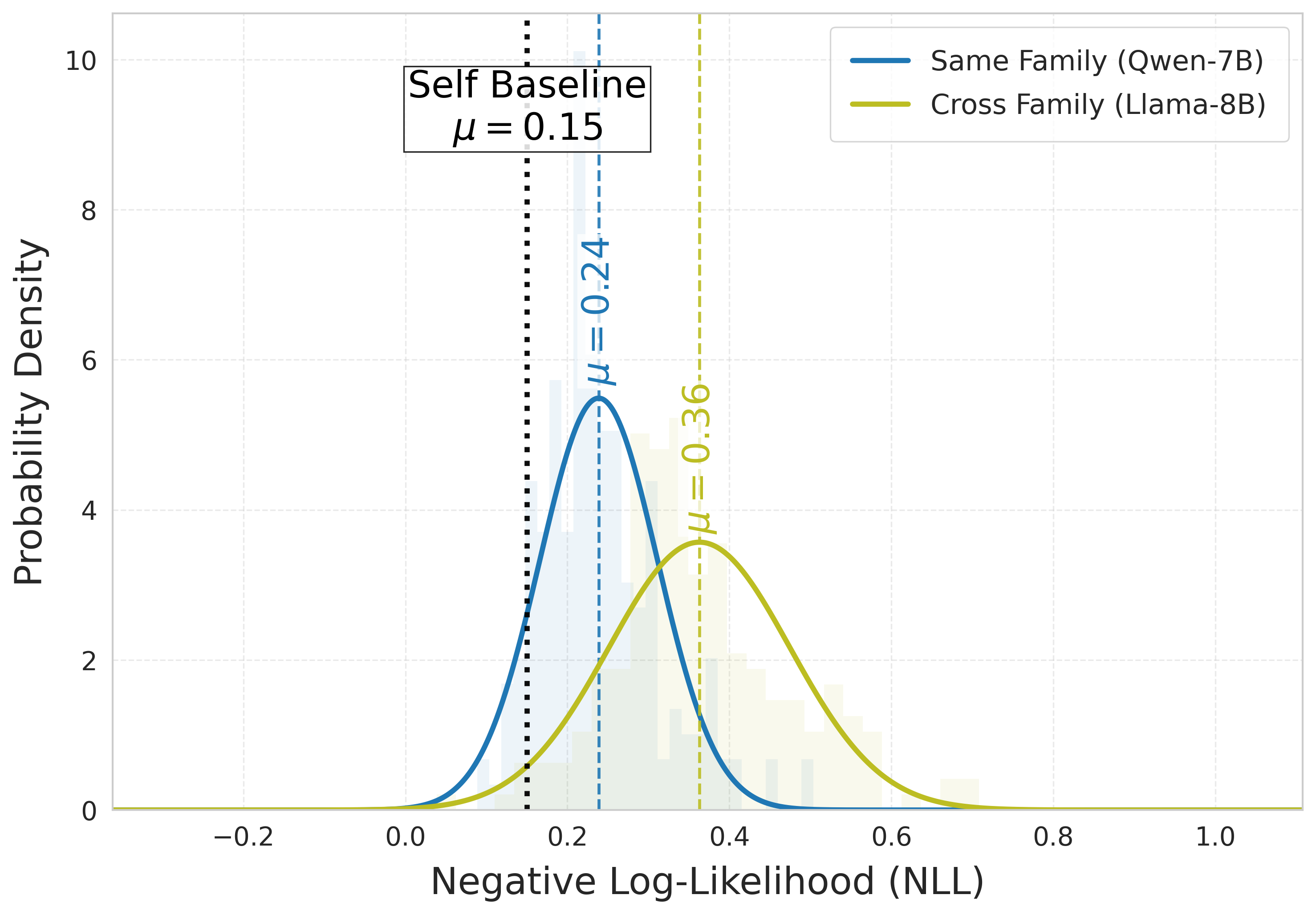}
        \caption{Same-family vs.\ cross-family reference samples evaluated under the target model.}
        \label{fig:nll_family_comparison_a}
    \end{subfigure}

    \vspace{0.6em}

    \begin{subfigure}[t]{0.9\linewidth}
        \centering
        \includegraphics[width=\linewidth]{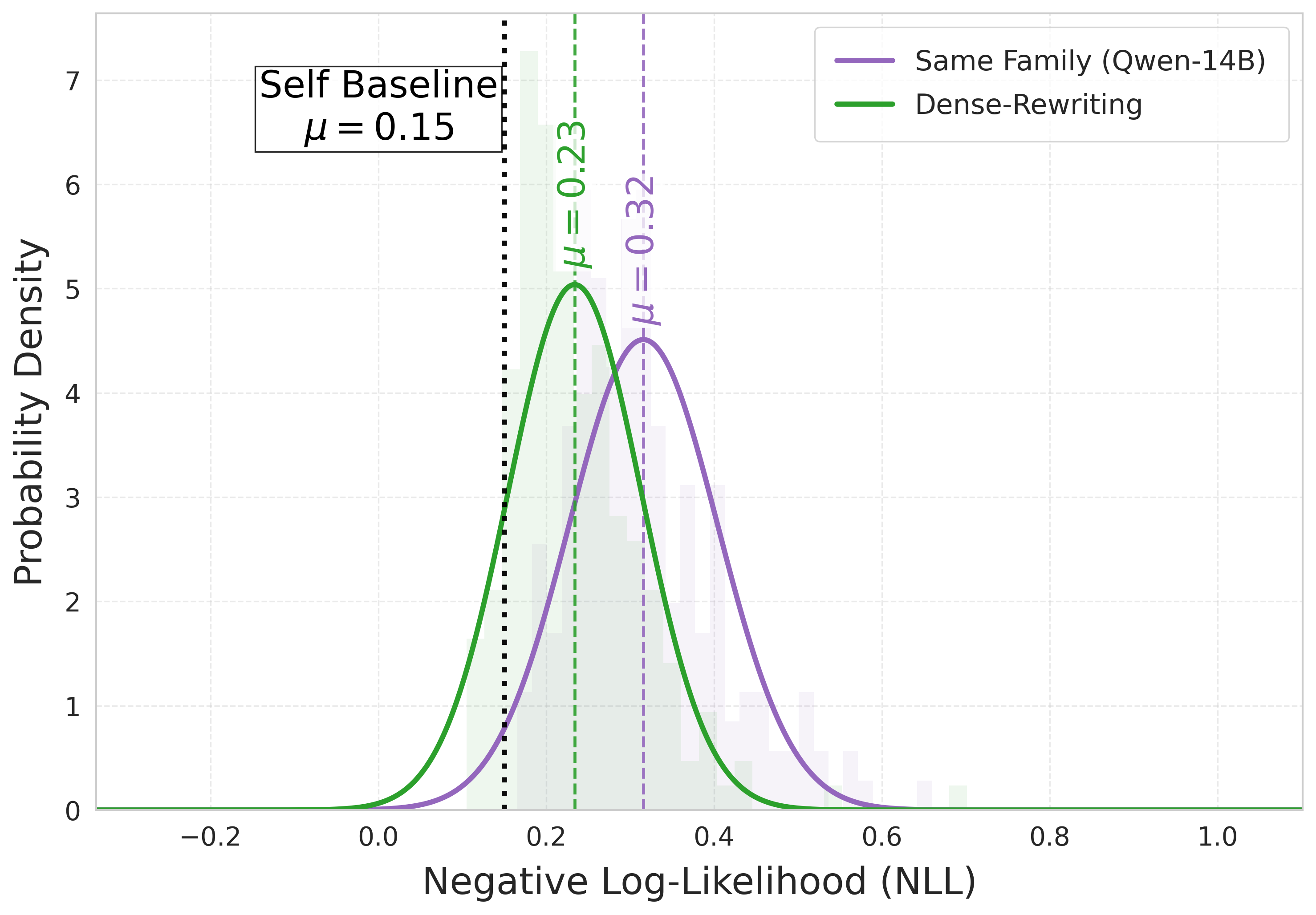}
        \caption{Same-family references vs.\ \rewrite target-model responses.}
        \label{fig:self_rewrite_nll_b}
    \end{subfigure}

    \caption{
    Distribution of token-level NLL evaluated under Qwen2.5-3B-Instruct on 200 GSM8K samples.
    (\textbf{a}) Reference reasoning traces generated by same-family (Qwen2.5-7B-Instruct) and cross-family (Llama-3.2-8B-Instruct) models.
    (\textbf{b}) Comparison between the same-family reference traces and \rewrite target-model responses.
    The black dotted line indicates the target model’s self-likelihood baseline.
    }
    \label{fig:nll_comparison}
\end{figure}

As shown in Figure~\ref{fig:nll_family_comparison_a}, reference samples from different model families exhibit substantially different NLL when evaluated under the same target model. In particular, cross-family references consistently incur higher NLL, indicating a larger distributional discrepancy from the target model’s intrinsic generation prior.

\subsection{From Alien Reference to Dense Self-Guidance}
\label{subsec:self_dense_guidance}
Building on the observation that the self-likelihood baseline is the lowest, we revisit a fundamental design choice for our inference-time steering: rather than relying on positive samples generated by external reference models, we ask whether effective positive samples can be constructed by \emph{minimally modifying the target model’s own responses}, thereby preserving distributional alignment while improving reasoning quality.

This idea is further reinforced by our analysis of the Qwen-2.5 model family (Figure~\ref{fig:stats_comparison}), where we observe that stronger models consistently exhibit higher \textit{reasoning density}, indicating a systematic structural difference in how reasoning is organized. We identify \reasoning phenomenon where more accurate mathematical reasoning is characterized by the following:
\begin{itemize}
    \item \textbf{Fewer total steps}, reducing the number of intermediate ``jump points'' where logical errors may accumulate,
    \item \textbf{Higher information density per step ($\rho$)}, ensuring that sufficient computation and contextual information are carried out within each step before transitioning.
\end{itemize}
Here, we formally define \textit{Reasoning Density} ($\rho$) as
\begin{equation}
    \text{\emph{Reasoning Density}, } \rho = \frac{N_{\text{tokens}}}{N_{\text{steps}}},
\end{equation}
where $N_{\text{tokens}}$ denotes the total sequence length and $N_{\text{steps}}$ is the number of reasoning steps, delimited by double newlines (\texttt{\textbackslash n\textbackslash n}).

These findings collectively motivate our \ours framework: to improve reasoning quality, one should steer the model not towards an alien reference, but towards a mathematically denser version of itself. Based on this, our goal is to rewrite the SLM's output from their default ``sparse'' reasoning style (low $\rho$) towards a ``dense'' style (high $\rho$) as a positive sample, which we refer to as \rewrite. 

Remarkably, these \rewrite responses achieve lower NLL than same-family reference traces, often approaching the target model’s self-likelihood baseline (Figure~\ref{fig:self_rewrite_nll_b}).
This result indicates that rewrite-based positive samples preserve distributional compatibility while inducing a shift toward denser reasoning. Together, these findings confirm that effective inference-time steering should operate within the target model’s intrinsic generation prior, guiding it toward improved reasoning structure.

\section{\ours}
\label{method}

\subsection{Steering Vector Extraction via Mean Difference}
\label{sec:vector_extraction}
With our preliminary analysis, we introduce \ours, a lightweight steering paradigm that shifts the target model’s hidden representations toward a subspace associated with \emph{dense reasoning}. Our framework derives steering directions from the target model itself via rewrite-based contrastive pairs, which is \rewrite.
An overview of the framework is illustrated in Figure~\ref{fig:03_method_overview}.

We begin by constructing a small calibration set consisting of $N$ contrastive pairs
$\mathcal{D} = \{(x_{\text{pos}}^{(i)}, x_{\text{neg}}^{(i)})\}_{i=1}^{N}$,
where $x_{\text{pos}}^{(i)}$ denotes the reasoning trace by \rewrite and $x_{\text{neg}}^{(i)}$ denotes the corresponding original (sparse) reasoning trace produced by the target model. Each pair preserves the same semantic content and differs primarily in reasoning structure. The detailed prompts we used for \rewrite are provided in Appendix~\ref{app:rewrite_prompt}.

We denote the output of the transformer block at layer $\ell$ as the \emph{residual stream} of layer $\ell$, and use $H_\ell(x) \in \mathbb{R}^{T \times d}$ to represent the sequence of hidden states when an input sequence $x$ of length $T$ is fed into the model, where $d$ is the hidden dimension.
With a slight abuse of notation, we denote $h_{\ell,t}(x)$ as the hidden state at token position $t$ in layer $\ell$.

For each reasoning trace, we summarize its representation using the activation corresponding to the \emph{final token} of the sequence.
This choice provides a compact, sequence-level summary while avoiding alignment issues arising from variable sequence lengths between dense and sparse reasoning variants.

Formally, the steering vector at layer $\ell$ is defined as the average contrastive difference between dense and sparse samples:
\begin{equation}
    v_\ell
    =
    \frac{1}{N}
    \sum_{i=1}^{N}
    \left(
    h_\ell(x_{\text{pos}}^{(i)})[-1]
    -
    h_\ell(x_{\text{neg}}^{(i)})[-1]
    \right),
\end{equation}
where $h_\ell(x)[-1]$ denotes the residual-stream activation associated with the final token of sequence $x$ at layer $\ell$.

The resulting vector $v_\ell$ captures a dominant direction in the activation space that separates dense reasoning from sparse reasoning.
By construction, this direction reflects structural differences in reasoning while remaining closely aligned with the target model’s intrinsic generation prior.

\subsection{Inference-Time Steering}
\label{sec:inference_steering}

During the inference phase, we perform a lightweight intervention to guide the SLM. For a new query $q$, at each decoding step $t$, we inject the pre-computed steering vector $v_\ell$ into the residual stream at a selected layer $\ell^\ast$ during autoregressive decoding.

The modified hidden state $\tilde{h}_{l,t}$ is computed as:
\begin{equation}
    \tilde{h}_{\ell^\ast,t}
    =
    h_{\ell^\ast,t}
    +
    \lambda \cdot v_{\ell^\ast},
\end{equation}
where $\lambda \in \mathbb{R}$ is a scalar hyperparameter controlling the strength of the steering intervention.

Positive values of $\lambda$ bias the generation toward the dense-reasoning subspace encoded by $v_{\ell^\ast}$, encouraging fewer but more information-rich reasoning steps.
When $\lambda$ is too large, the intervention may distort the target model’s intrinsic generation dynamics and degrade output quality.
Conversely, when $\lambda$ is close to zero, the steering effect becomes negligible.
In our experiments, we identify a stable range of $\lambda$ that preserves fluency while consistently improving reasoning structure.

\begin{table*}[t]
\small
\centering
\caption{\textbf{Results on Mathematical Reasoning Benchmarks.}
We compare Zero-shot CoT prompting, prompt engineering, knowledge distillation,
\infam, and \ours.
\textbf{Bold} indicates the best performance. Avg. is sample-weighted over all evaluation questions. Results for smaller models and SEAL are provided in Appendix~\ref{app:additional_results}.}
\label{tab:main_results}
 \resizebox{\textwidth}{!}{  
\begin{tabular}{l l l c c c c c c}
\toprule
Target Model & &  Method & GSM8K & MATH500 & AMC & Olympiad Bench & AIME & Avg. \\
\midrule

\multirow{8}{*}{Qwen-2.5-3B-Instruct}
 & \multirow{2}{*}{\textit{Prompt-based Baselines}} 
 &  Zero-shot & 83.6 & 63.0 & \textbf{42.5} & 20.0 & 0.0 & 61.2 \\
&  &  Prompt Engineering & 20.0 & 32.8 & 30.0 & 9.8 & 6.7 & 19.8 \\
\cmidrule(lr){2-9}
 & \multirow{4}{*}{\textit{Distillation Methods}} 
 &  Short CoT & 79.9 & 58.6 & 30.0 & 18.1 & 6.7 & 57.8 \\
& &  Long CoT  & 82.5 & 49.8 & 25.0 & 12.7 & 0.0 & 55.9 \\
& &  Mix-Long  & 82.2 & 60.4 & 30.0 & 20.6 & 6.7 & 60.0 \\
& &  Mix-Large & 83.7 & 61.6 & 37.5 & \textbf{21.0} & 6.7 & 61.3 \\

\cmidrule(lr){2-9}
 & \multirow{2}{*}{\textit{Ours}} 
 &  \cellcolor{gray!15}\infam  & \cellcolor{gray!15}\textbf{85.3} & \cellcolor{gray!15}59.8 & \cellcolor{gray!15}37.5 & \cellcolor{gray!15}20.0 & \cellcolor{gray!15}0.0 & \cellcolor{gray!15}61.4 \\
& &  \cellcolor{gray!15}\textbf{\ours} &\cellcolor{gray!15} 84.8 &\cellcolor{gray!15} \textbf{64.6} & \cellcolor{gray!15}\textbf{42.5} & \cellcolor{gray!15}20.7 & \cellcolor{gray!15}\textbf{10.0} & \cellcolor{gray!15}\textbf{62.5} \\
\midrule

\multirow{8}{*}{Llama-3.2-3B-Instruct}
 & \multirow{2}{*}{\textit{Prompt-based Baselines}} 
 &  Zero-shot & 79.0 & 46.0 & 15.0 & 9.78 & \textbf{10.0} & 52.5 \\
 & &  Prompt Engineering & 31.3 & 19.6 & 20.0 & 5.3 & 3.3 & 21.7 \\
\cmidrule(lr){2-9}
 & \multirow{4}{*}{\textit{Distillation Methods}} 
 &  Short CoT & 77.4 & 48.8 & 20.0 & \textbf{12.7} & 6.7 & 53.1 \\
 & &  Long CoT  & 77.3 & 40.6 & 17.5 & 9.2 & 3.3 & 50.4 \\
& &  Mix-Long  & 78.7 & \textbf{51.6} & \textbf{25.0} & 12.3 & 3.3 & \textbf{54.2} \\
& &  Mix-Large & 77.0 & 46.6 & 20.0 & 11.6 & 3.3 & 52.1 \\
\cmidrule(lr){2-9}
 & \multirow{2}{*}{\textit{Ours}} 
 &  \cellcolor{gray!15}\infam & \cellcolor{gray!15}\textbf{80.9} & \cellcolor{gray!15}44.4 & \cellcolor{gray!15}\textbf{25.0} & \cellcolor{gray!15}10.5 &\cellcolor{gray!15} 6.7 & \cellcolor{gray!15}53.5 \\
& &  \cellcolor{gray!15}\textbf{\ours} & \cellcolor{gray!15}79.7 & \cellcolor{gray!15}43.4 & \cellcolor{gray!15}\textbf{25.0} & \cellcolor{gray!15}10.5 & \cellcolor{gray!15}6.7 & \cellcolor{gray!15}52.7 \\

\bottomrule
\end{tabular}
}
\vspace{0.3em}
\end{table*}
\section{Experiments}
\label{exps}

\subsection{Experimental Setup}

\paragraph{Datasets.}
We use GSM8K \cite{cobbe2021training} as the primary benchmark for mathematical reasoning.
Unless otherwise specified, steering vectors are constructed using a held-out subset of 50 GSM8K problems that do not overlap with the evaluation set.
To assess robustness on more challenging, out-of-domain (OOD) problems, we further evaluate on \textbf{MATH-500} \citep{hendrycks2021measuringmathematicalproblemsolving},
\textbf{AIME 2024}, \textbf{AMC 2023}, and the English math subset of \textbf{OlympiadBench} \citep{he2024olympiadbenchchallengingbenchmarkpromoting}.

\paragraph{Method Variants.}
We evaluate both variants of our method:
(i) \ours, our steering by \rewrite and
(ii) \infam, which derives steering directions from the same family but a larger counterpart's responses as positive samples.

\paragraph{Models}
To evaluate the scalability and universality of our method across different architectures and scales, we conduct experiments on two model families: Qwen-2.5 instruct (1.5B, 3B, 7B), and Llama-3.1/3.2 (1B, 3B, 8B).
For each family, the largest model (7B for Qwen and 8B for Llama) is additionally used to generate reference reasoning traces when constructing steering vectors in the \infam setting.

\paragraph{Implementation Details.}
All inference experiments use greedy decoding to ensure deterministic and reproducible results. The maximum generation length is set to 2048 tokens to accommodate long chain-of-thought sequences.
For \ours, we extract steering vectors using $N=50$ contrastive pairs constructed via the method described in Section \ref{subsec:self_dense_guidance}. For \infam, we perform a hyperparameter sweep over both the steering layer and coefficient $\lambda$, searching $\lambda \in [-20, 20]$ across a set of candidate layers on a held-out validation set (a subset of GSM8K training set) to find the appropriate $\lambda$ for our target models.

\subsection{Baselines}
We compare \ours against three categories of baselines to establish its effectiveness:

\begin{enumerate}
    \item \textbf{CoT Prompting (Zero-shot):} We evaluate the performance of the vanilla model using the standard prompt described in Appendix~\ref{app:cot_prompt}.
    \item \textbf{Prompt Engineering (Dense Style):} We apply an inference-time version of our \rewrite prompt (Appendix~\ref{app:inference_prompt}) directly during generation without steering. This examines whether the performance gain comes solely from the prompt instructions or from the internal activation steering.
    \item \textbf{Knowledge Distillation (Training-based):} We compare against state-of-the-art fine-tuned models to benchmark our inference-time intervention against training-based methods. These include:
    \begin{itemize}
        \item \textbf{Short / Long CoT.} \cite{li2025small}
            Models fine-tuned on curated short or long chain-of-thought (CoT) datasets.
            Short CoT consists of concise reasoning traces with fewer steps and shorter token lengths,
            while Long CoT contains extended, reflective reasoning sequences generated by stronger teacher models.
            
        \item \textbf{Mix-Long / Mix-Large.} \cite{li2025small}
            Training-based distillation baselines that mix different sources of reasoning traces.
            Mix-Long combines long and short CoT data with a fixed ratio,
            while Mix-Large blends reasoning traces generated by large and small teacher models.
            These methods aim to reduce distributional mismatch and improve the learnability of small models.
    \end{itemize}  
    \item \textbf{Reasoning Steering Baseline:} We compare against SEAL \citep{chen2025seal}, a training-free steering method that calibrates long-CoT reasoning by steering away from redundant reflection and transition patterns.
\end{enumerate}

\subsection{Main Results}
Overall, our results demonstrate that \ours provides a robust inference-time alternative to training-based distillation methods, achieving consistent gains across model families and evaluation benchmarks.
\paragraph{Performance on GSM8K.} 
Table \ref{tab:main_results} presents the comparative results for GSM8K. Our \ours consistently outperforms the Zero-shot CoT baseline across all model scales. For instance, on Qwen-2.5-3B-Instruct, we observe a significant improvement over the distillation method from 82.2\% to \textbf{84.8\%} and \infam even performs better. A similar trend is observed on Llama-3.2-3B-Instruct. These results provide two important insights.
First, the comparable performance of \ours and \infam, which exhibit similar negative log-likelihood (NLL), suggests that our NLL-based selection strategy is effective for identifying high-quality steering positive samples.
Second, the consistent improvements across both variants demonstrate that steering internal representations can reliably enhance multi-step reasoning performance when evaluated on the in-domain dataset.

\paragraph{Performance on Other Math Datasets.}
Beyond GSM8K, we evaluate all methods on more challenging and out-of-domain mathematical benchmarks, including MATH500, AMC, OlympiadBench, and AIME.
As shown in Table~\ref{tab:main_results}, our steering-based methods consistently surpass prompt-based baselines and remain competitive with, or outperform, distillation-based approaches across these datasets. The results provide us with two insights.
First, our strong performance on out-of-domain benchmarks indicates that the learned steering vectors capture transferable reasoning patterns rather than overfitting to surface-level prompt cues.
This suggests that the contrastive pairs used for steering encode meaningful mathematical reasoning signals that can generalize across various problem distributions. Second, \ours achieves particularly strong results on \emph{harder} benchmarks such as AMC and AIME.
For example, on Qwen-2.5-3B-Instruct, \ours attains the best performance on AIME (10.0\%) and matches or exceeds the strongest baselines on AMC.
These outperformances indicate that inference-time steering is especially effective for tasks requiring deeper, multi-step reasoning, where small models typically struggle the most.

\paragraph{Overall Performance.}
Because no method dominates every individual benchmark, we also report a sample-weighted average across GSM8K, MATH-500, AMC, OlympiadBench, and AIME in Table~\ref{tab:main_results}.
On Qwen2.5-3B-Instruct, our approaches achieve the best overall performance in almost all methods.
On Llama-3.2-3B-Instruct, the results are more mixed, but our approaches remain competitive with the strongest distillation baselines.
These results support our main claim that dense steering is a lightweight and competitive inference-time alternative.

\begin{AIbox}[title=Takeaway 1]
    \makecell{\ours induces denser reasoning patterns in small models, enabling stronger and more robust multi-step reasoning.}
\end{AIbox}


\begin{figure}[t]
    \centering
    \includegraphics[width=0.85\linewidth]{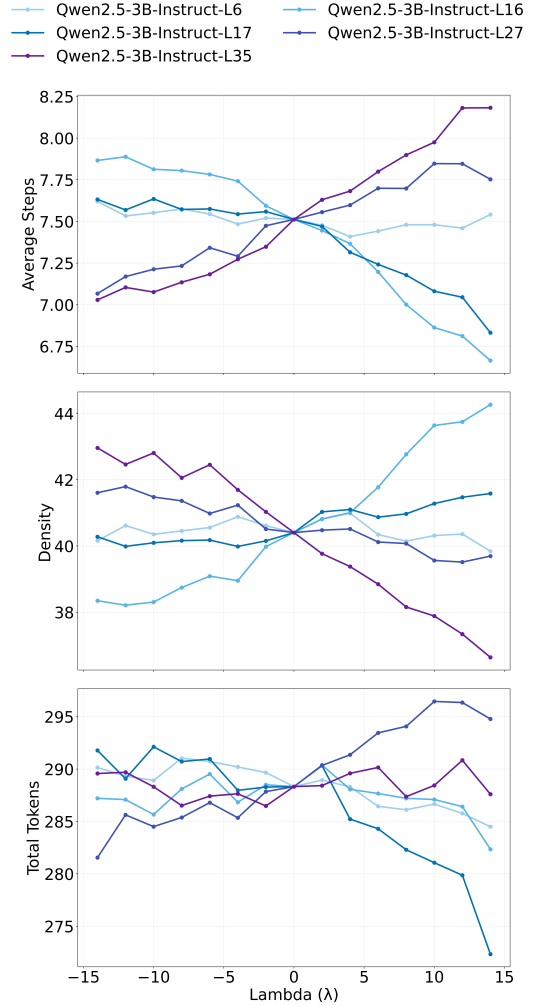}
    \caption{Sensitivity analysis on Qwen2.5-3B-Instruct.
   Metric trends for reasoning steps (top), density (middle), and total tokens (bottom) are shown across varying $\lambda$. 
    Middle layers (L16/L17) exhibit the most effective and stable control over reasoning length, while early layers show limited sensitivity, and later layers tend to be unstable.
    }
    \label{fig:sensitivity_analysis}
\end{figure}

\subsection{Prompt Engineering vs. Steering}
We further compare our steering-based approach with prompt engineering using the same rewrite-style instructions. Empirically, prompt engineering performs poorly across most benchmarks. Upon inspecting the generated outputs, we observe that the complex prompt used in Appendix~\ref{app:inference_prompt} often causes the model to directly produce final answers without explicit intermediate reasoning steps.
This behavior suggests that small language models, in general, frequently fail to correctly interpret or follow complex reasoning-style instructions, leading to instruction-following failures or collapsed chain-of-thought generation.
Upon inspecting the generated outputs, we observe that prompt engineering often causes the model to directly produce final answers without explicit intermediate reasoning steps.
In contrast, steering counts on an external, reliable reference distribution.
In \infam, the reference is provided by a larger model from the same family, while in \ours, it is derived from high-quality rewrites produced by a stronger commercial model.
These references provide the steering vectors with reliable positive signals, enabling small models to align their internal representations toward effective reasoning trajectories.
As a result, steering bypasses the limitations of textual instruction following and directly enforces the desired reasoning behavior at the representation level.
This explains why steering-based methods remain effective even when prompt engineering fails, especially for small-scale models.
\begin{AIbox}[title=Takeaway 2]
    \makecell{\ours enables reliable knowledge transfer from large models to small models.}
\end{AIbox}

\subsection{Knowledge Distillation vs. Steering}

As shown in Table~\ref{tab:main_results}, \ours and \infam achieve competitive and often improved performance compared to distillation-based methods, while operating entirely at inference time without requiring additional training or teacher-generated datasets.
This difference can be attributed to the fact that steering modulates internal representations directly during inference, which may help mitigate optimization instability and distribution mismatch commonly observed in distillation for small models.

In terms of data efficiency, distillation-based methods typically rely on thousands of teacher-generated samples (over 2,000 in our comparison) and additional training stages.
In contrast, our steering approach leverages significantly fewer contrastive pairs (50 in our experiments) to guide reasoning behavior at inference time.

\begin{AIbox}[title=Takeaway 3]
    \makecell{\ours provides a more efficient and stable alternative to training-based knowledge distillation for improving small-model reasoning.}
\end{AIbox}

\section{Analysis}
\label{sec:analysis}

\subsection{Layer and Lambda Sensitivity}
\label{sec:ablation_layer_lambda}

To understand how activation steering influences the reasoning dynamics, we conducted a sensitivity analysis across different layers ($L \in \{6, 16, 17, 27, 35\}$) and steering intensities ($\lambda \in [-14, 14]$). As shown in Figure~\ref{fig:sensitivity_analysis}, the model's response to steering varies significantly by depth.

\textbf{Layer Sensitivity.} We observe a distinct functional division across the network depth:
\begin{itemize}
    \item \textbf{Early Layers (e.g., L6):} These layers exhibit minimal sensitivity to the steering vector. The curves for steps and token counts remain relatively flat regardless of $\lambda$. This exhibits that early layers are contributing to learn primitive features rather than dense reasoning patterns.
    \item \textbf{Middle Layers (e.g., L16, L17):} These layers display the highest responsiveness \cite{templeton2024scaling}. Specifically, Layer 17 shows a dramatic reduction in both \textit{Average Steps} and \textit{Total Tokens} as $\lambda$ increases. That is, steering at this depth effectively makes the reasoning chain denser, thereby enhancing model performance, which we aim for.
    \item \textbf{Late Layers (e.g., L27, L35):} Late-layer intervention yields divergent behaviors. For instance, L35 (red) actually increases the number of steps and total tokens as $\lambda$ increases, opposing the trend seen in L17. This suggests that late-layer representations are already close to specific output tokens, and intervening here conflicts with the established reasoning trajectory, leading to compensatory or verbose generation.
\end{itemize}

\begin{AIbox}[title=Takeaway 4]
    \makecell{Middle layers act as the primary control point for \textbf{Dense Reasoning}.}
\end{AIbox}

\begin{figure}[t]
    \centering
    \includegraphics[width=0.8\linewidth]{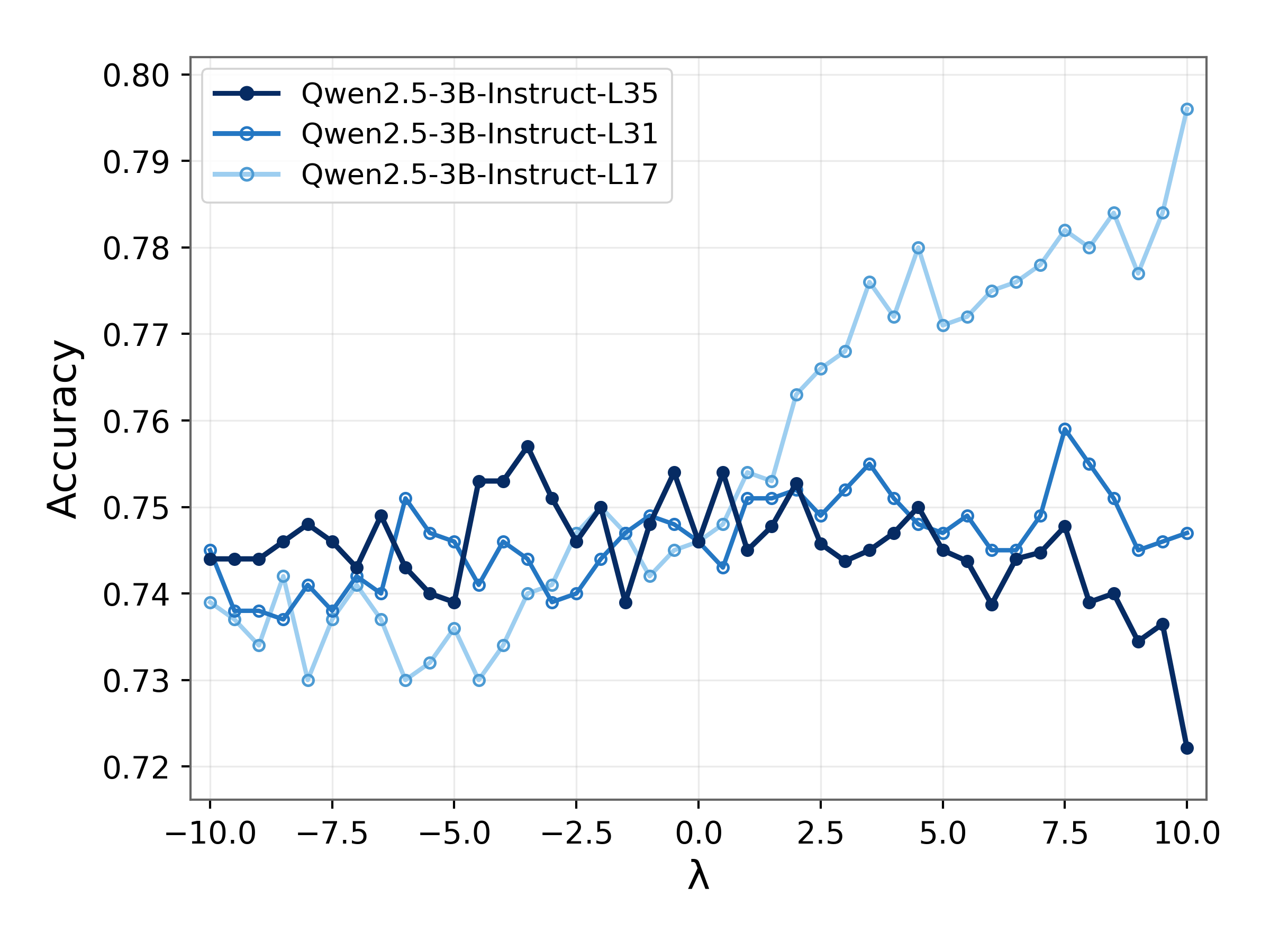}
    \caption{\textbf{Impact of Steering Strength on Accuracy.} Reasoning accuracy on GSM8K improves for Layer 17 (middle layer) as the steering coefficient $\lambda$ increases. This demonstrates that steering toward the dense reasoning effectively enhances model performance.}
    \label{fig:acc_lambda}
\end{figure}

\begin{figure}[t]
    \centering
    \includegraphics[width=0.8\linewidth]{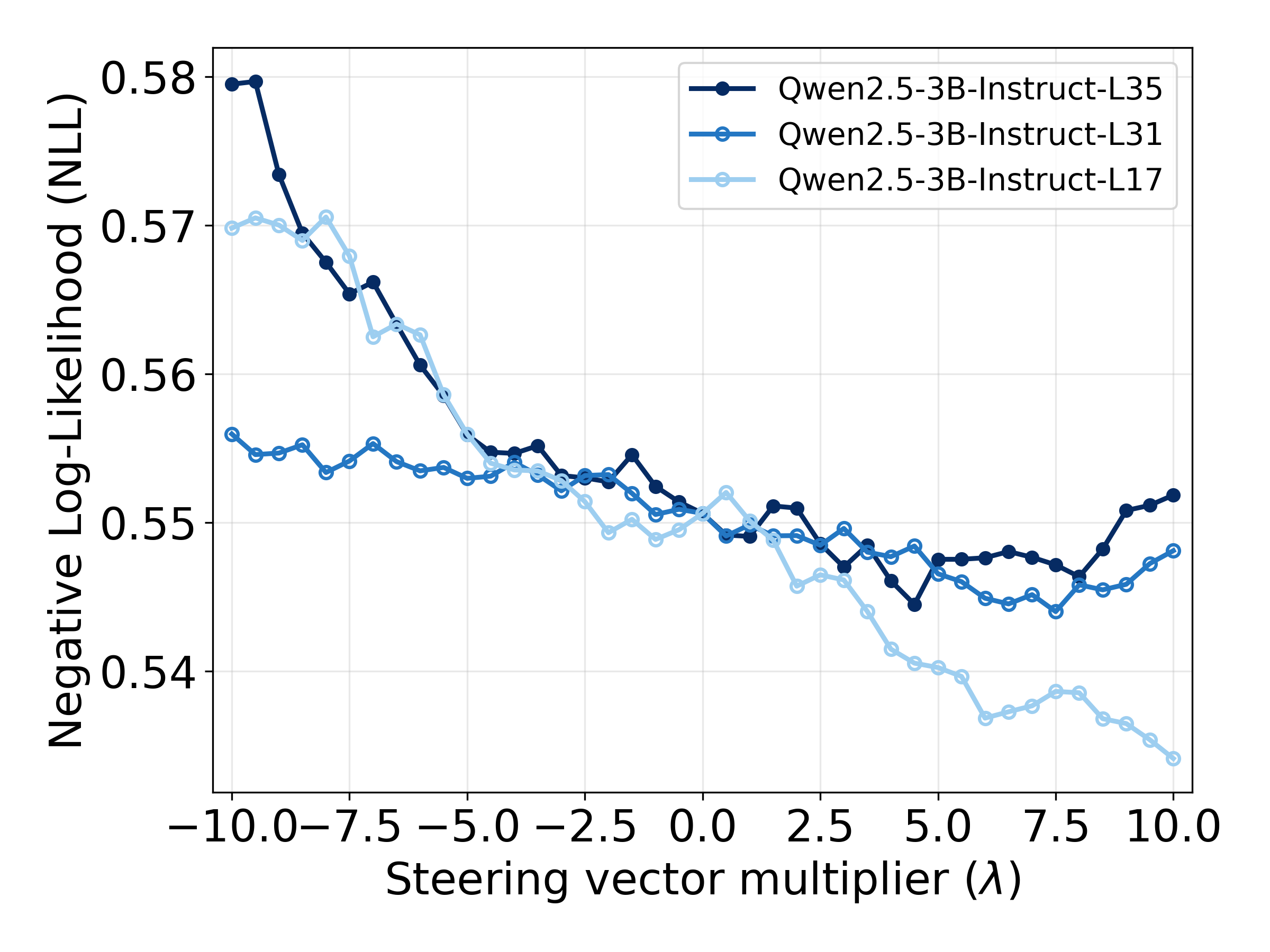}
    \caption{\textbf{Impact of Steering Strength on Token-Level NLL.} Average token-level negative log-likelihood (NLL) of steered reasoning traces under the base model as a function of $\lambda$. Steering towards dense reasoning reduces NLL, indicating improved alignment with the model’s intrinsic generation prior.}
    \label{fig:ppl_lambda}
\end{figure}

\textbf{Lambda Sensitivity.} Beyond steering location, the magnitude of $\lambda$ also plays a critical role in steering effectiveness. In Figure \ref{fig:sensitivity_analysis}, moderate values (e.g., $\lambda \in [0, 10]$ for L17) successfully guide the model towards the target reasoning style. However, pushing $\lambda$ beyond this range often leads to excessive verbosity, as evidenced by the sharp increase in variance for total tokens at $\lambda > 10$.

To further understand how steering strength affects both model confidence and generation behavior, we analyze the resulting changes in token-level NLL, accuracy, and token dynamics as $\lambda$ varies.

Figure \ref{fig:acc_lambda} shows the overall task accuracy under different steering intensities.
Consistent with our layer sensitivity analysis, steering at the intermediate layer (L17) yields a monotonic improvement in accuracy as $\lambda$ increases, while late-layer steering (L35) becomes unstable and degrades performance at larger $\lambda$.
This indicates that moderate dense steering at appropriate depths effectively improves reasoning correctness for small models, whereas excessive or late-stage interventions conflict with the model’s established generation process.

Figure \ref{fig:ppl_lambda} examines how steering strength influences the model’s intrinsic confidence by measuring the token-level NLL of generated reasoning traces under the base model.
We observe that effective steering (e.g., Layer 17 with appropriate $\lambda$) consistently reduces NLL, suggesting that the steered trajectories are not only more accurate but also more aligned with the model’s natural generation prior.

Taken altogether, these results reveal that $\lambda$ plays a critical role in steering:
moderate steering improves both accuracy and NLL by guiding the model toward high-confidence, dense reasoning trajectories while excessive steering destabilizes generation and undermines performance.

\subsection{Out-of-Domain Transfer}
\label{addition_results}
To examine the generalizability and transferability of \ours, we evaluate the same pipeline on LogiQA~\cite{liu2020logiqa}, a logical reasoning benchmark.
As shown in Table~\ref{tab:logiqa_results}, both \ours and \infam substantially outperform the zero-shot baseline, improving exact-match accuracy from 44.22\% to 58.22\% and 60.22\%, respectively.
This suggests that dense, model-aligned reasoning is not limited to math reasoning but can transfer to broader logical reasoning tasks.

\begin{table}[h]
\centering
\caption{\textbf{LogiQA transfer results.} EM stands for exact-match accuracy.}
\label{tab:logiqa_results}
\begin{tabular}{lccc}
\toprule
Method & Best Layer & Best $\lambda$ & EM (\%) \\
\midrule
Baseline & -- & -- & 44.22 \\
\ours & L16 & -4.0 & 58.22 \\
\infam & L2 & -1.5 & 60.22 \\
\bottomrule
\end{tabular}
\end{table}

We also evaluate broader out-of-domain robustness on MMLU~\citep{hendrycks2020measuring}, BBH CoT~\citep{suzgun2023challenging}, and HotpotQA~\cite{yang-etal-2018-hotpotqa}.
As shown in Table~\ref{tab:ood_results}, the steering methods remain comparable to the baseline and do not exhibit obvious degradation.
\begin{table}[h]
\centering
\caption{\textbf{Out-of-domain robustness.} HotpotQA is reported with F1; the other tasks use accuracy.}
\label{tab:ood_results}
\begin{tabular}{lccc}
\toprule
Task & Baseline & \ours & \infam \\
\midrule
MMLU & 64.61 & 64.62 & 62.74 \\
BBH CoT & 54.42 & 54.05 & 54.26 \\
HotpotQA & 46.64 & 45.87 & 47.62 \\
\bottomrule
\end{tabular}
\end{table}

\section{Conclusion}
We presented \ours, a method to enhance SLMs' reasoning by steering internal representations toward a denser one. By reducing the number of reasoning steps while increasing reasoning density, we enhance the performance on math benchmarks. Our results on multiple benchmarks show that inference-time steering is a powerful, low-cost approach to fine-tuning for improving small model reasoning.

\section*{Impact Statement}
This paper presents work whose goal is to advance the research in the field of Machine Learning, especially reasoning proficiency. There are no potential societal consequences of our work that we feel must be specifically highlighted here.

\section*{Limitations}
\ours is an inference-time structural intervention rather than a method for adding new knowledge to the base model.
It can reorganize reasoning behavior that is already latent in the model, but it cannot reliably solve problems that require missing facts, unavailable skills, or substantially stronger search.
Finally, this insight is mostly validated on mathematical reasoning, with additional transfer checks on logical and multi-hop reasoning; broader domains, larger model families, and more mechanistic attribution analyses remain important future work.

\bibliography{references}
\bibliographystyle{icml2026}

\newpage
\appendix
\onecolumn
\section*{Appendix}

\section{Prompts details}
\label{app:prompts}

\subsection{Baseline: Chain-of-Thought Prompt}
\label{app:cot_prompt}

\begin{promptbox}{Normal Chain-of-Thought Prompt}
    Solve the following math problem. Present the final answer in the format: 
    Final Answer: \\boxed{your_answer}.\nProblem: {{problem}}\nAnswer:
\end{promptbox}

\subsection{Rewrite Prompt for Positive Sample Construction}
\label{app:rewrite_prompt}

Below is the full prompt template used to generate the ``dense'' positive samples ($x_{\text{pos}}$). We apply this prompt to rewrite the granular chain-of-thought into a denser format using GPT 5.1.

\begin{promptbox}{Dense-Rewriting Prompt}
You will lightly rewrite the solution by CONSERVATIVELY merging steps, 
while keeping the SAME style and meaning.

Hard constraints:
- Keep the SAME meaning and do NOT change the final conclusion/answer 
  implied by the solution.
- Do NOT invent new reasoning. Only compress/merge/rephrase existing steps.
- Keep the style and tone the SAME as the original (do not change formality, 
  phrasing habits, or formatting conventions).
- Only merge steps when it is NECESSARY and safe (e.g., two adjacent lines 
  that are clearly redundant or tightly coupled).
  Do NOT aggressively minimize the number of steps. If merging would change 
  the “feel” or clarity, keep the original steps.
- When you merge, prefer merging 2 adjacent steps into 1 step (avoid merging 
  many lines at once).
- Keep computations consistent with the original (same numbers/operations, 
  no new math).
- Preserve special markers like "<<a=b>>" if they appear; do not introduce 
  many new ones.
- Output plain text only. No bullet points or added commentary.

Question:
{question}

Original solution (model output):
{original_resp}

Now output ONLY the rewritten solution (same style, with a few necessary merges):
\end{promptbox}

\subsection{Baseline: Dense Reasoning Prompt at Inference Time}
\label{app:inference_prompt}

\begin{promptbox}{Dense Reasoning Prompt at Inference Time}
You are generating a solution by lightly and conservatively merging reasoning steps.
You must strictly follow these rules:
- Preserve the exact meaning of the original reasoning and the final conclusion.
- Do not introduce any new reasoning, assumptions, or computations.
- Do not remove necessary reasoning steps; only merge or lightly rephrase existing ones.
- Keep the style, tone, and formatting conventions identical to the original solution.
- Merge steps only when it is clearly safe, typically between two adjacent and tightly coupled steps.
- Do not aggressively reduce the number of steps.
- When merging, prefer merging at most two adjacent steps at a time.
- Keep all numbers, operations, and results exactly the same.
- Preserve any special markers such as "<<a=b>>" if they appear.
- Output plain text only, with no explanations, commentary, or meta-text.
\end{promptbox}

\section{Additional Results}
\label{app:additional_results}
\subsection{Qwen2.5-1.5B-Instruct and Llama3.2-1B-Instruct Results}
Table~\ref{tab:appendix_results} reports results on smaller models.
These models are substantially weaker than the 3B models in the main experiments, so the gains from inference-time steering are more limited and less uniform.
Nevertheless, \ours remains competitive with distillation-based baselines and improves over zero-shot on most benchmarks.

\begin{table*}[h]
\small
\centering
\caption{\textbf{Results on Mathematical Reasoning Benchmarks.}
We compare Zero-shot CoT prompting, prompt engineering, knowledge distillation,
\infam, and \ours.
\textbf{Bold} indicates the best performance. Avg. is sample-weighted over all evaluation questions.}
\label{tab:appendix_results}
\resizebox{\textwidth}{!}{  
\begin{tabular}{l l l c c c c c c}
\toprule
Target Model & & Method & GSM8K & MATH500 & AMC & Olympiad Bench & AIME & Avg. \\
\midrule
\multirow{6}{*}{Llama-3.2-1B-Instruct}
 & \multirow{2}{*}{\textit{Prompt-based Baselines}} 
 & Zero-shot & 43.1 & 18.4 & 2.5 & 2.2 & \textbf{3.3} & 24.0 \\
& & Prompt Engineering & 3.9 & 1.0 & 2.5 & 2.1 & \textbf{3.3} & 2.7 \\
\cmidrule(lr){2-9}
 & \multirow{2}{*}{\textit{Distillation Methods}} 
 & Short CoT & \textbf{49.2} & \textbf{33.4} & \textbf{7.5} & \textbf{7.4} & 0.0 & \textbf{31.7} \\
& & Long CoT & 42.3 & 28.6 & 2.5 & 5.5 & 0.0 & 26.9 \\
\cmidrule(lr){2-9}
 & \multirow{2}{*}{\textit{Ours}} 
 & \cellcolor{gray!15}\infam & \cellcolor{gray!15}44.3 & \cellcolor{gray!15}14.2 & \cellcolor{gray!15}0.0 & \cellcolor{gray!15}2.8 & \cellcolor{gray!15}0.0 & \cellcolor{gray!15}23.7 \\
& & \cellcolor{gray!15}\textbf{\ours} & \cellcolor{gray!15}44.8 & \cellcolor{gray!15}15.6 & \cellcolor{gray!15}2.5 & \cellcolor{gray!15}4.1 & \cellcolor{gray!15}0.0 & \cellcolor{gray!15}24.7 \\

\midrule
 
\multirow{6}{*}{Qwen2.5-1.5B-Instruct}
 & \multirow{2}{*}{\textit{Prompt-based Baselines}} 
 & Zero-shot & 63.1 & 40.0 & 30.0 & 11.6 & 0.0 & 41.0 \\
& & Prompt Engineering & 42.4 & 32.6 & 30.0 & 9.3 & \textbf{3.3} & 29.5 \\
\cmidrule(lr){2-9}
 & \multirow{2}{*}{\textit{Distillation Methods}} 
 & Short CoT & 70.8 & \textbf{50.0} & 27.5 & \textbf{12.4} & \textbf{3.3} & \textbf{46.9} \\
& & Long CoT & \textbf{72.3} & 36.2 & 10.0 & 9.5 & 0.0 & 43.3 \\
\cmidrule(lr){2-9}
 & \multirow{2}{*}{\textit{Ours}} 
 & \cellcolor{gray!15}\infam & \cellcolor{gray!15}68.9 & \cellcolor{gray!15}45.2 & \cellcolor{gray!15}20.0 & \cellcolor{gray!15}11.6 & \cellcolor{gray!15}\textbf{3.3} & \cellcolor{gray!15}44.6 \\
& & \cellcolor{gray!15}\textbf{\ours} & \cellcolor{gray!15}65.1 & \cellcolor{gray!15}41.4 & \cellcolor{gray!15}\textbf{32.5} & \cellcolor{gray!15}10.8 & \cellcolor{gray!15}0.0 & \cellcolor{gray!15}42.0 \\

\bottomrule
\end{tabular}
}
\vspace{0.3em}
\end{table*}
\subsection{Comparison with SEAL}
Since Qwen2.5-3B-Instruct does not natively produce explicit long-reasoning traces such as \texttt{<think>} regions, we apply SEAL to operate over the full generated solution.
As shown in Table~\ref{tab:seal_baseline}, SEAL gives a small gain on GSM8K but underperforms the zero-shot baseline on MATH-500, while \ours improves both benchmarks.

\begin{table}[H]
\small
\centering
\caption{\textbf{Comparison with SEAL on Qwen2.5-3B-Instruct.}}
\label{tab:seal_baseline}
\begin{tabular}{lcc}
\toprule
Method & GSM8K & MATH-500 \\
\midrule
Zero-shot & 83.6 & 63.0 \\
SEAL & 83.8 & 61.2 \\
\ours & \textbf{84.8} & \textbf{64.6} \\
\bottomrule
\end{tabular}
\end{table}

\section{Controls for Rewriting and Compression}
\label{app:control_rewriting_compression}

We conduct two control groups to distinguish semantically correct dense reasoning from superficial compression and from possible hidden knowledge transfer by the external rewriter.
All methods are evaluated on the same GSM8K control run with Qwen2.5-3B-Instruct.

\paragraph{Controls for superficial density.}
We first test whether the gain can be explained by arbitrary rewriting or shortening.
\textit{Random paraphrase} rewrites the original solution without explicitly increasing density.
\textit{Random step compression} randomly merges local reasoning steps.
\textit{Dense-but-incorrect traces} uses compact traces whose reasoning or final answer is incorrect.
As shown in Table~\ref{tab:control_acc}, all three controls improve slightly over the baseline but remain below \ours.
This suggests that density alone is insufficient: the dense trace must also preserve correct reasoning content.

\paragraph{Controls for rewriter dependence.}
We next test whether the gain mainly comes from GPT-5.1 injecting new reasoning ability.
\textit{Rule-based rewriting} uses deterministic local step-merging rules without an LLM rewriter.
\textit{GPT-5-mini rewriting} replaces GPT-5.1 with a weaker rewriter under the same rewriting constraints.
\textit{Direct GPT-5.1 answers} uses answers generated directly by GPT-5.1 instead of conservative rewrites of the target model's own traces.
The rule-based and GPT-5-mini controls still improve over the baseline, but underperform \ours, indicating that closed-source GPT-5.1 is not essential but provides higher-quality contrastive pairs.
Direct GPT-5.1 answers are also weaker than \ours, suggesting that the benefit does not come from directly transferring GPT-5.1 knowledge.

\begin{table}[h]
\centering
\caption{\textbf{Control directions on GSM8K with Qwen2.5-3B-Instruct.}}
\label{tab:control_acc}
\begin{tabular}{lc}
\toprule
Method & Accuracy (\%) \\
\midrule
Baseline & 83.8 \\
\midrule
Random paraphrase & 84.6 \\
Random step compression & 84.3 \\
Dense-but-incorrect traces & 84.3 \\
\midrule
Rule-based rewriting & 84.4 \\
GPT-5-mini rewriting & 84.7 \\
Direct GPT-5.1 answers & 84.1 \\
\midrule
\ours & \textbf{85.9} \\
\bottomrule
\end{tabular}
\end{table}

\begin{figure}[h]
\centering
\includegraphics[width=0.65\linewidth]{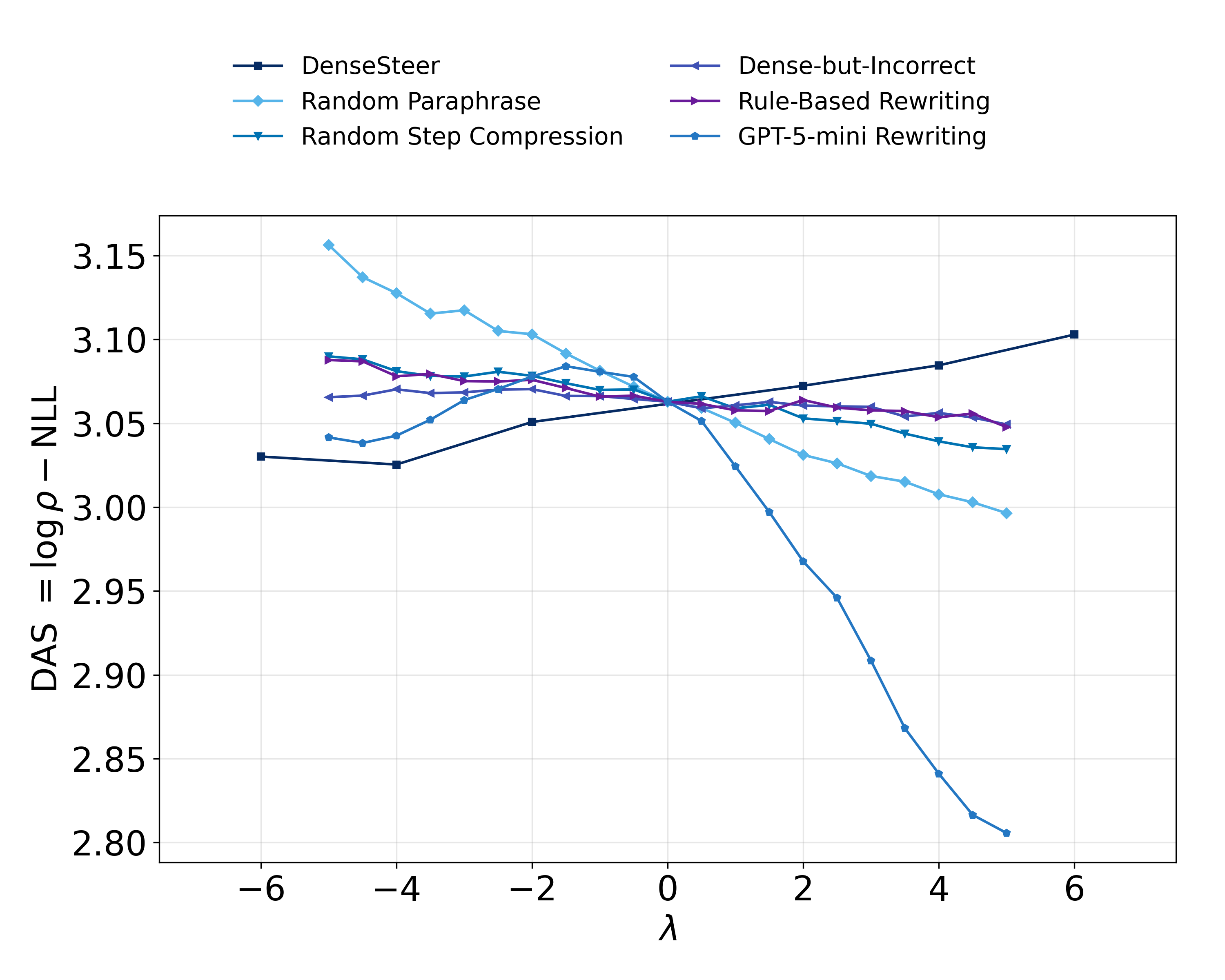}
\caption{\textbf{DAS under steering controls.}}
\label{fig:das_controls}
\end{figure}

\paragraph{Density-alignment analysis.}
To further quantify whether steering improves density without sacrificing model compatibility, we define a diagnostic \textit{Density-Alignment Score} (DAS):
\begin{equation}
    \mathrm{DAS}(y;q) = \log \rho(y) - \mathrm{NLL}(y;P_{\theta}),
\label{eq:das_appendix}
\end{equation}
where $\rho(y)$ is the reasoning density of trace $y$ and $\mathrm{NLL}(y;P_{\theta})$ is its token-level negative log-likelihood under the target model.
Higher DAS indicates traces that are denser while remaining more compatible with the target model distribution.
As shown in Figure~\ref{fig:das_controls}, \ours is the only direction whose DAS consistently increases with positive steering, while the control directions stay flat or decline.
Together, these controls suggest that the benefit comes from semantically correct, model-aligned dense reasoning rather than arbitrary compression or external rewriting alone.

\section{Rewrite Audit}
\label{app:rewrite_audit}

\begin{figure}[h]
\centering
\includegraphics[width=1.0\linewidth]{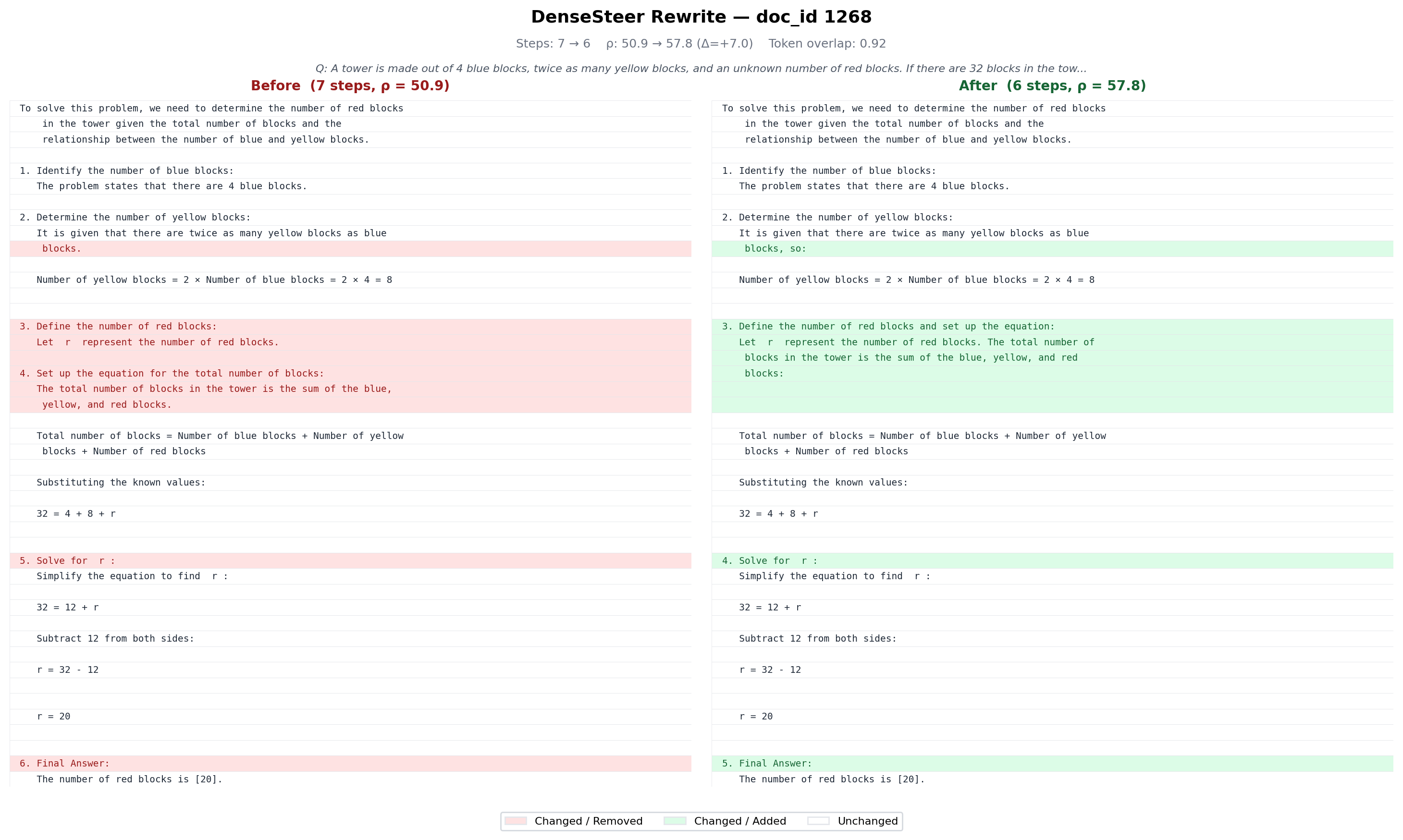}
\caption{\textbf{Qualitative rewrite audit.}
The dense rewrite primarily merges adjacent redundant steps while preserving the original computations and final answer.}
\label{fig:rewrite_sample}
\end{figure}

To check whether the external rewriter injects new reasoning, we compare DenseSteer rewrites with two alternatives: a deterministic rule-based step-merging procedure and GPT-5-mini rewriting.
As shown in Table~\ref{tab:rewrite_audit}, GPT-5.1 produces a moderate rewrite: it changes the original trace more than the nearly identity-preserving rule-based method, but is less aggressive than GPT-5-mini.
The qualitative example in Figure~\ref{fig:rewrite_sample} further shows that the rewrite mainly merges adjacent redundant steps while preserving computations and the final answer.
An LLM-as-a-judge audit and manual inspection found that all three rewriting strategies preserve the final answer and original reasoning, and fix 0 errors in the calibration pairs.
This suggests that GPT-5.1 acts primarily as a high-quality structural rewriter rather than as a source of hidden reasoning transfer.

\begin{table}[h]
\centering
\caption{\textbf{Rewrite audit for positive-pair construction.}
Negative samples have 7.02 steps and a density of 38.78 on average.}
\label{tab:rewrite_audit}
\begin{tabular}{lccc}
\toprule
Metric & \ours & Rule-Based & GPT-5-mini \\
\midrule
\multicolumn{4}{l}{\textit{Rewrite behavior}} \\
Steps (rewrite) & 6.76 & 6.60 & 5.84 \\
Density (rewrite) & 38.40 & 41.14 & 39.85 \\
Edit similarity & 88.0\% & 99.9\% & 74.6\% \\
Adjacent merge ratio & 10.8\% & 10.7\% & 42.3\% \\
\midrule
\multicolumn{4}{l}{\textit{Correctness audit}} \\
Final answer preserved & 100\% & 100\% & 100\% \\
Original reasoning preserved & 100\% & 100\% & 100\% \\
Errors fixed by rewriter & 0 & 0 & 0 \\
\midrule
\multicolumn{4}{l}{\textit{Steered outputs}} \\
Steps (steered) & 7.03 & 7.14 & 7.41 \\
Density (steered) & \textbf{40.94} & 40.19 & 38.24 \\
\bottomrule
\end{tabular}
\end{table}

\section{Calibration Set Size}
\begin{figure}[h]
\centering
\includegraphics[width=0.5\linewidth]{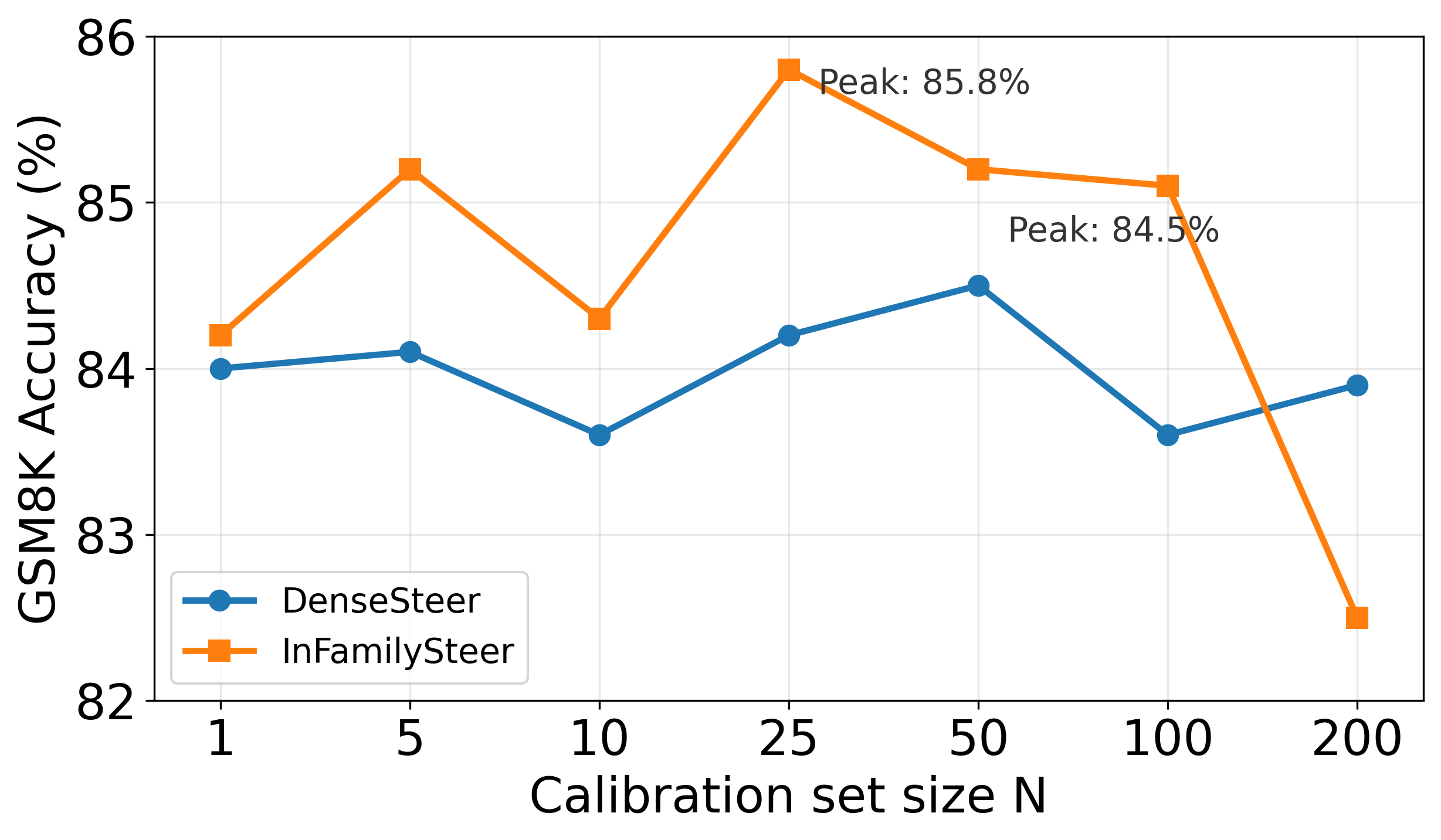}
\caption{\textbf{Calibration set-size ablation.} \ours and \infam reach strong performance with few contrastive pairs and remain most effective around 25--50 pairs.}
\label{fig:cali_ablation}
\end{figure}

Figure~\ref{fig:cali_ablation} shows that DenseSteer is effective even with very small calibration sets and most effective around 25--50 pairs.
Larger calibration sets do not necessarily improve performance, suggesting that the dense-reasoning direction is strong but can be diluted by heterogeneous examples.

\end{document}